\documentclass[lettersize,journal, journal, twoside]{IEEEtran}
\usepackage{amsmath,amsfonts}
\usepackage{algorithmic}
\usepackage{array}
\usepackage[caption=false,font=normalsize,labelfont=sf,textfont=sf]{subfig}
\usepackage{textcomp}
\usepackage{stfloats}
\usepackage{url}
\usepackage{verbatim}
\usepackage{graphicx}
\def\BibTeX{{\rm B\kern-.05em{\sc i\kern-.025em b}\kern-.08em
    T\kern-.1667em\lower.7ex\hbox{E}\kern-.125emX}}
\usepackage{balance}
\newcommand{\etal}{et al.}
\usepackage{hyperref}
\usepackage{amsmath}
\usepackage{tabularx}
\usepackage{booktabs}
\usepackage{caption}
\usepackage[ruled,linesnumbered]{algorithm2e}
\usepackage{xcolor}

\newcommand{\reflabel}{dummy} 

\newcommand{\be}{\begin{equation}}
\newcommand{\ee}{\end{equation}}
\newcommand{\eqlabel}[1]{\label{eq:\reflabel-#1}}
\renewcommand{\eqref}[2][\reflabel]{(\ref{eq:#1-#2})}

\newcommand{\seclabel}[1]{\label{sec:\reflabel-#1}}
\newcommand{\secref}[2][\reflabel]{Section~\ref{sec:#1-#2}}

\newcommand{\figlabel}[2][\reflabel]{\label{fig:#1-#2}}
\newcommand{\figref}[2][\reflabel]{Fig.~\ref{fig:#1-#2}}

\newcommand{\tablelabel}[2][\reflabel]{\label{table:#1-#2}}
\newcommand{\tableref}[2][\reflabel]{Table~\ref{table:#1-#2}}

\let\oldtwocolumn\twocolumn
\renewcommand\twocolumn[1][]{%
    \oldtwocolumn[{#1}{
    \begin{center}
    \includegraphics[trim=0 1 0 2, clip, width=0.9\textwidth, height=0.3\linewidth]{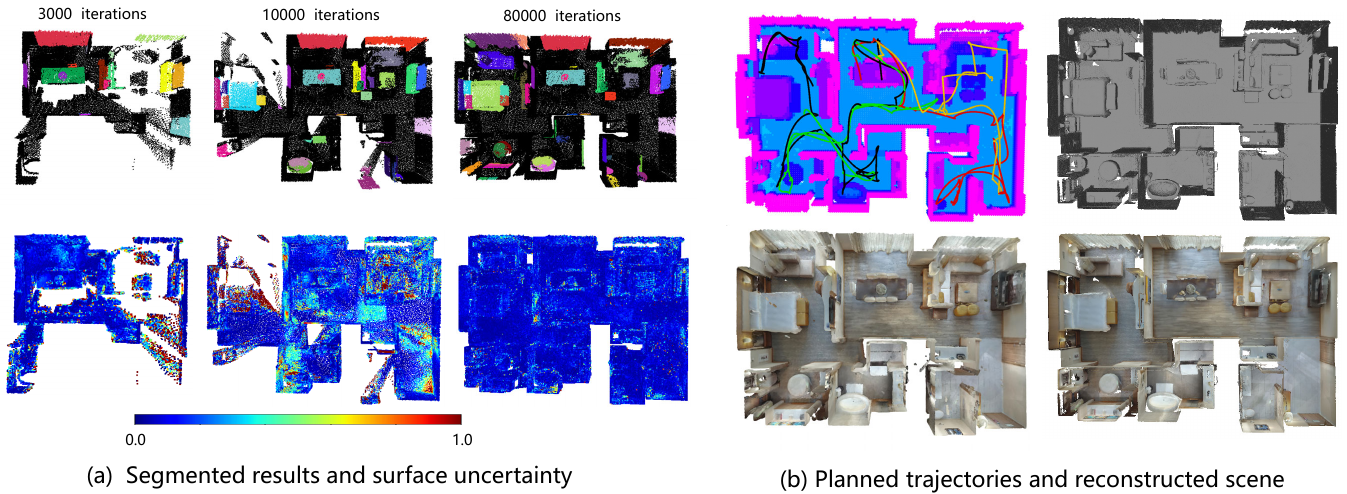}
      \captionof{figure}{(a) Segmented results (Top) and Surface uncertainty (Bottom) from reconstruction iterations; (b) Planned trajectories (Top left, four robots); Reconstructed scene: 3DGS (Bottom left), mesh with texture (Bottom right) and without texture (Top right) from ASH.}
    \label{teaser}
    \end{center}
    }
    ]
}

\begin{document}

\title{Multi-robot autonomous 3D reconstruction using Gaussian splatting with Semantic guidance}
\author{Jing Zeng$^{1}$, Qi Ye$^{1}$$^{*}$, Tianle Liu$^{1}$, Yang Xu$^{1}$, Jin Li$^{2}$, Jinming Xu$^{1}$, Liang Li$^{1}$, Jiming Chen$^{1}$  
\thanks{$^{1}$ College of Control Science and Engineering, Zhejiang University, Hangzhou, 310027, China.}
\thanks{$^{2}$ College of Information Engineering, Zhejiang University of Technology, Hangzhou, 310023, China.}
\thanks{$^{*}$ Qi Ye (Corresponding author, qi.ye@zju.edu.cn) is with the College of Control Science and Engineering, the State Key Laboratory of Industrial Control Technology, Zhejiang University, and the Key Lab of CS\&AUS of Zhejiang Province.}
}
\maketitle

\thispagestyle{empty}


\begin{abstract}

Implicit neural representations and 3D Gaussian splatting (3DGS) have shown great potential for scene reconstruction. Recent studies have expanded their applications in autonomous reconstruction through task assignment methods. However, these methods are mainly limited to single robot, and rapid reconstruction of large-scale scenes remains challenging. Additionally, task-driven planning based on surface uncertainty is prone to being trapped in local optima. To this end, we propose the first 3DGS-based centralized multi-robot autonomous 3D reconstruction framework. 
To further reduce time cost of task generation and improve reconstruction quality, we integrate online open-vocabulary semantic segmentation with surface uncertainty of 3DGS, focusing view sampling on regions with high instance uncertainty.
Finally, we develop a multi-robot collaboration strategy with mode and task assignments improving reconstruction quality while ensuring planning efficiency. 
Our method demonstrates the highest reconstruction quality among all planning methods and superior planning efficiency compared to existing multi-robot methods. We deploy our method on multiple robots, and results show that it can effectively plan view paths and reconstruct scenes with high quality.

\end{abstract}


\section{Introduction}

\IEEEPARstart{R}{econstruction} of indoor scenes is essential for numerous applications, such as gaming, robotics, augmented and virtual reality~\cite{gu2023conceptgraphs,matsuki2024gaussian,jiang2023h}. 
Recently, implicit neural representations have demonstrated
significant potential in autonomous systems due to their precise
3D reconstruction quality~\cite{ran2023neurar,zeng2024autonomous}. 
To achieve higher rendering speed and quality, GS-Planner~\cite{jin2024gs} proposes the first active 3D reconstruction system using 3DGS with online evaluation. Despite the impressive results of these works, their primary limitation is the inherent design to operate with only one single robot, significantly reducing scanning efficiency in large indoor environments. 


 Multi-robot autonomous 3D reconstruction provides advantages over single-robot systems, including broader coverage and enhanced efficiency~\cite{guo2022asynchronous,hardouin2023multirobot}. To address the task assignment problem in multi-robot collaboration, a Multi-depot multi-traveling salesman problem (MDMTSP) needs to be solved. Considering that MDMTSP is NP-hard and lacks an efficient exact solution, some improved methods, such as cluster-and-assign~\cite{guo2022asynchronous} and greedy cluster assignment~\cite{hardouin2023multirobot}, have been employed to accelerate the process of robot mode and task assignment. Although these methods achieve relatively favorable results in mode and task assignment, they still require repeated iterations to approximate optimal assignment. When the number of tasks increases, repeated computations of TSP entail significant computational costs, reducing planning efficiency. 
 
However, during the generation of multiple tasks in active reconstruction methods~\cite{lauri2020multi,dong2019multi,hardouin2023multirobot}, either the coverage of frontiers~\cite{hardouin2023multirobot} or the coverage of specified surface uncertainty regions~\cite{lauri2020multi,dong2019multi} is prioritized, leading to objects not being reconstructed with high quality due to rapid scanning without specific attention. Specifically, when defined or learned uncertainties contain noise, excessive focus on high uncertainty areas can cause the robot to move back and forth in large scenes, potentially getting trapped in local optima. To address this problem, semantic methods~\cite{liu2018object,guo2022asynchronous} are adopted to make viewpoint sampling more focused. Liu~\etal~\cite{liu2018object} proposed a model-driven objectness metric to evaluate the similarity and completeness of segmented components extracted from objects in a 3D model database. Moreover, Guo~\etal~\cite{guo2022asynchronous} utilized incompleteness scores of segmented objects derived from point cloud completion to inform and optimize task generation. Nevertheless, building a 3D model database~\cite{liu2018object} or performing point cloud completion~\cite{guo2022asynchronous} requires significant human and time resources.



To handle the first challenge, we propose a novel centralized framework for multi-robot autonomous 3D reconstruction using 3DGS. To address the second challenge, we propose an efficient multi-robot collaboration strategy incorporating global-local tasks for robot mode assignment and an improved K-means algorithm for task assignment. By separating mode assignment from task assignment and focusing task clustering solely on Euclidean distance instead of TSP distance, this method eliminates repetitive TSP calculations typically required for iterative assignment in MDMTSP. To tackle the third challenge, we integrate online open-vocabulary semantic segmentation with 3DGS surface uncertainty, prioritizing view sampling in regions with high instance uncertainty to reduce redundant view generation, which eliminates the need for significant effort in pre-establishing a model database and avoids the time cost associated with point cloud completion. 

To summarize, our contributions are:
\begin{itemize}
    \item We propose the first centralized multi-robot autonomous 3D reconstruction framework utilizing 3DGS, which comprises three modules: Perception, Task generation, and Hierarchical planning.
    \item We incorporate online open-vocabulary semantic segmentation and surface uncertainty of 3DGS, focusing view sampling on areas with high instance uncertainty, which reduces the generation of redundant viewpoints, thereby reducing time for task generation.
    \item We propose an efficient multi-robot collaboration strategy: global-local tasks for mode assignment of robots and improved K-means for task assignment, improving reconstruction quality while ensuring planning efficiency.
\end{itemize}

\section{Related work}

\subsection{Autonomous 3D reconstruction}
Neural radiance field (NeRF) has become a highly effective method for 3D scene reconstruction due to its remarkable ability to render photorealistic images~\cite{mildenhall2020nerf} and represent scene geometry~\cite{yu2022monosdf} effectively. However, achieving high visual quality still necessitates using neural networks, which are expensive to train and render. More recently, 3DGS~\cite{kerbl20233d} has demonstrated comparable or superior rendering performance to NeRF~\cite{mildenhall2020nerf}, achieving faster rendering and optimization speeds with an order of magnitude.

View path planning is essential in autonomous reconstruction, focusing on optimizing the sequence of viewpoints to reconstruct a 3D scene efficiently. Ran~\etal~\cite{ran2023neurar} propose the first autonomous 3D reconstruction system using an implicit neural representation. To enhance global coverage capabilities in complex environments while avoiding local minima, Zeng~\etal~\cite{zeng2024autonomous} incorporate frontier-based exploration tasks with surface-based reconstruction tasks to improve the efficacy of reconstruction. Jin~\etal~\cite{jin2024gs} leverages the advantageous features of 3DGS to incorporate real-time quality and completeness assessment to guide the robot's reconstruction process. 

\subsection{Multiple-robot collaboration}
Recently, multi-task assignment methods~\cite{lauri2020multi,dong2019multi,hardouin2023multirobot} have been utilized in multi-robot active reconstruction to ensure efficient global coverage, precise scene reconstruction, and load balancing among multiple robots. Lauria~\etal~\cite{lauri2020multi} considers a similar matroid-constrained submodular maximization problem for multisensor NBV planning. Dong~\etal~\cite{dong2019multi} formulate task assignments based on Optimal mass transport (OMT) and propose efficient solutions based on a divide-and-conquer scheme. Hardouin~\etal~\cite{hardouin2023multirobot} introduce multiagent NBV planners to route robots to viewpoint configurations.

\subsection{Semantic guidance in planning}
Due to the focused nature of semantic information and corresponding segmentation attributes for each object in the scene, some methods~\cite{liu2018object,zheng2019active,guo2022asynchronous} have attempted to use semantic information to guide robots in scanning the scene.
Liu~\etal~\cite{liu2018object} proposes a model-driven objectness metric to measure the similarity and completeness of segmented components from objects in the 3D model database. Zheng~\etal~\cite{zheng2019active}estimate the next best view based on the uncertainty in scene reconstruction and understanding. Furthermore, Guo~\etal~\cite{guo2022asynchronous} introduces semantic information into the multi-robot system, guiding task generation through incompleteness score based on point cloud completion.

\section{Method}

\subsection{Problem Statement and System Overview} 
In this paper, we address the challenge of multi-robot autonomous scene reconstruction with 3DGS utilizing a centralized framework. This study aims to achieve efficient exploration and scanning of an unknown indoor scene. By starting from the initial positions of multiple robots, the goal is to maximize scanning coverage and reconstruction quality while minimizing scanning effort~\cite{dong2019multi, guo2022asynchronous}.

\begin{figure*}
\centering
\subfloat[]{
		\includegraphics[width=0.23\linewidth, height=0.16\linewidth]{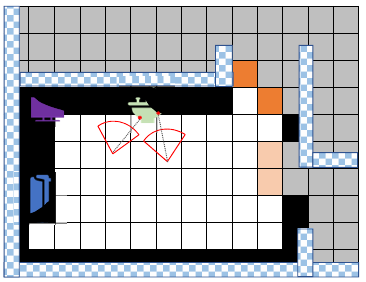}}
\subfloat[]{
		\includegraphics[width=0.23\linewidth, height=0.16\linewidth]{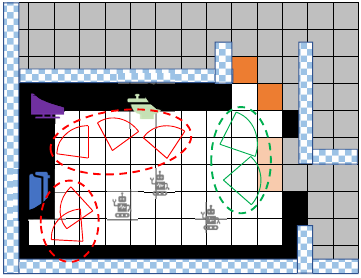}}
\subfloat[]{
		\includegraphics[width=0.23\linewidth, height=0.16\linewidth]{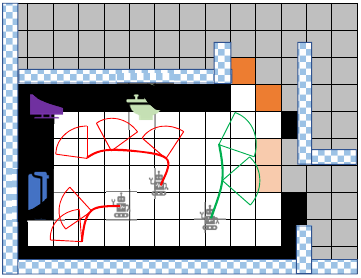}}
\subfloat{
		\includegraphics[width=0.25\linewidth, height=0.16\linewidth]{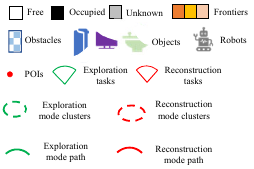}}
\caption{Overview of our method. (a) Semantic-guided reconstruction task generation. (b) Multi-robot collaboration for mode and task assignments. (c) Single-robot view path planning.}
\label{method_details}
\vspace{-5mm}
\end{figure*}

Under the framework of the centralized strategy, our pipeline comprises three components, as illustrated in~\figref{pipeline}. The method overview is shown in Fig.~\ref{method_details}. \textbf{The Perception module} receives images rendered from the Unity at specified viewpoints similar to~\cite{ran2023neurar, zeng2023efficient} and maintains four representations (\secref{perception}). A volumetric representation (occupancy grid map) is adopted for exploration tasks. 3DGS~\cite{yugay2023gaussian}  and  Open-vocabulary 3D instance retrieval (OVIR-3D)~\cite{lu2023ovir} are adopted for reconstruction tasks. A modern framework for Parallel spatial hashing (ASH)~\cite{dong2022ash} is used for scene geometry acquisition and noise point pruning of 3DGS. In \textbf{the Task generation module} (\secref{task_generation}), surface uncertainty is obtained through 3DGS, and reconstruction instances are acquired through OVIR-3D segmentation. These processes allow for determining uncertainty for each point in every segmented instance, thereby guiding the generation of reconstruction tasks. Exploration tasks are generated from frontiers extracted from the volumetric map~\cite{zhou2021fuel}. In \textbf{the Hierarchical planning Module} (\secref{hierarchical_planning}), Firstly, we determine the number of robots assigned to exploration and reconstruction modes based on the ratio between global whole exploration and reconstruction tasks. Each robot's mode is determined based on the number of tasks within its local range. Subsequently, for each robot, a global viewpoint path is planned using the Asymmetric traveling salesman problem (ATSP) solver, and B-spline trajectory optimization~\cite{tordesillas2019faster} is used to refine the path for smooth navigation. The images captured along this path are then processed by the perception system until the entire autonomous reconstruction process is completed.

\begin{figure*}[htbp]
    \centering
    \includegraphics[width=0.9\linewidth]{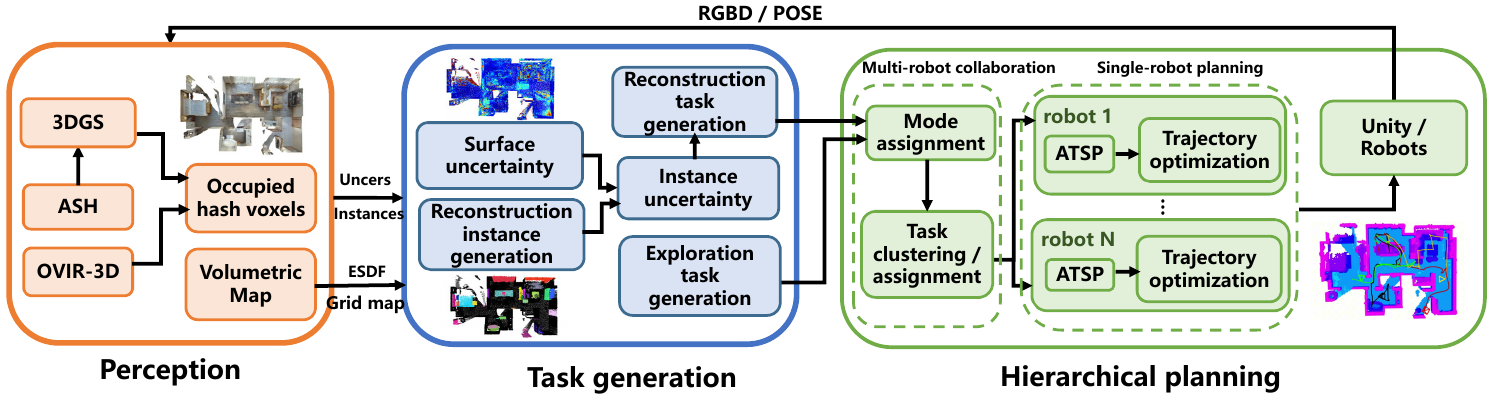}
    \caption{The pipeline of our proposed method.} 
    \figlabel{pipeline}
    \vspace{-5mm}
\end{figure*}

\subsection{Perception}
\seclabel{perception}
In this section, we first introduce the 3DGS representation and design a surface uncertainty retrieval method to evaluate its reconstruction quality. Then, we introduce an ASH)~\cite{dong2022ash} method to obtain scene geometry and prune noisy Gaussians. Finally, we employ open-vocabulary online 3D semantic segmentation to focus view sampling on instances, thereby reducing the generation of redundant viewpoints.

\subsubsection{3D Gaussian splatting with surface uncertainty}
In this study, we use Gaussian-SLAM~\cite{yugay2023gaussian} with online sub-map optimization as the base framework for our research. However, in our autonomous reconstruction system, the merging and quality evaluation of sub-maps require considerable computation time. Therefore, we maintain only a global 3DGS map during training, accelerating the quality evaluation process.

Our goal is to optimize a scene representation that captures the appearance of the scene, leading to a detailed, dense map and high-quality novel view synthesis. To achieve this, we represent the scene as a set of 3D Gaussians $\mathbf{G}^g$ as follows:
\begin{equation}
\mathbf{G}^g = \{ G^g_i : (\mu_i, {SC}_i, R_i, o_i, {SH}_i) \mid i = 1, \ldots, N \}
\eqlabel{3dgs}
\end{equation}

where each Gaussian $G^g_i$ is defined by a mean $\mu_i \in R^3$, scale ${SC}_i \in R^3$, rotation $R_i \in R^4$, opacity $o_i \in R$ and spherical harmonics ${SH}_i \in R^{48}$.



To evaluate the reconstruction quality of 3DGS, we adopt a loss caching strategy similar to GS-Planner~\cite{jin2024gs}. This process requires projecting the loss $L_{proj}$ from the image space to the world space and caching the loss into occupied voxels. Specifically, $L_{proj}$ is a weighted sum of color loss $L_{color}$ and depth loss $L_{depth}$:
\begin{equation}
L_{proj} = L_{color} + \lambda_d \cdot L_{depth}
\eqlabel{proj}
\end{equation}

where $\lambda_d = 0.5$ is the weight coefficient. To reduce computational overhead and memory usage, we maintain occupied hash voxels $H_{occ}$ with resolution 0.05 m to store the occupied voxels, which are updated by adding new images. The losses will be averaged for each projected voxel $v$ if it is already in $H_{occ}$. Otherwise, the new voxel with the position of projecting point $p$ and associated loss $L_p$ is added to $H_{occ}$. We then obtain uniform surface points 
$S$ from $H_{occ}$ and and surface uncertainty $U_{gs}$ with uncertainty $L_p$ of each point $p \in S$:


\subsubsection{ASH for scene geometry and pruning noisy Gaussians} It is widely recognized that 3DGS has achieved impressive results in novel view synthesis. However, for complex scenes, such as multi-room indoor environments, the geometric representation capability of 3DGS is still limited. Therefore, we introduce an ASH~\cite{dong2022ash} method to represent scene geometry. ASH implements spatially varying operations, transitioning seamlessly from volumetric geometry reconstruction to differentiable appearance reconstruction. ASH compensates for the limitations of 3DGS and can also prune noisy Gaussians generated during the training process of 3DGS. Specifically, ASH extracts geometric surface points at intervals of 30 frames and generates a KD-Tree. For each Gaussian in 3DGS, the nearest point in surface points of ASH is found, and if the distance is greater than 0.05 m, the Gaussian is considered noisy and is pruned. 

\subsubsection{Open-vocabulary online 3D semantic segmentation}
As the goal is to obtain high-quality reconstruction of 3D scenes, more scans should be conducted around surface points with high uncertainty, similar to AIISRFE~\cite{zeng2024autonomous}. However, due to the scattered distribution of surface uncertainty $U_{gs}$, excessive redundant views are generated in regions with high uncertainty points, thus reducing the scanning efficiency. 


OVIR-3D~\cite{lu2023ovir} aims to return a set of 3D instance segments given a 3D point cloud and the corresponding RGBD images with poses. This is achieved by a multi-view fusion of text-aligned 2D region proposals~\cite{zhou2022detecting, kirillov2023segment} into 3D space. To obtain the aforementioned 3D point cloud online and align it with the surface uncertainty of 3DGS, we employ surface points $S$ from the occupied voxel 
$H_{occ}$ as the online point cloud input for OVIR-3D segmentation. The 2D region proposals of each frame are then projected to the surface points $S$ given the camera pose. The projected 3D regions are either matched to existing 3D object instances $O = \{O_1, \ldots, O_b\}$ with 3D features $ F^{3D} = \{f^{3D}_1, \ldots, f^{3D}_b\}$, or added as a new instance if not matched with anything. Instance points of all objects are $ S_O = \{S_1, \ldots, S_b\}, \ S_O \in S$, where $b$ is the number of segmented 3D instances. Viewpoints are generated for incomplete object instances to overcome the local minima caused by the scattered distribution of surface uncertainty.


\vspace{-2mm}
\begin{algorithm}[htb]\footnotesize
\caption{Semantic-guided Reconstruction Task Generation}
\label{alg:rec}
\SetKwInOut{Input}{Input}
\SetKwInOut{Output}{Output}

\Input{Downsampling ratio $N_{down}$, POI distance threshold $d_{POI}$, 
       3DGS surface uncertainty $U_{gs}$, segmented instances $O$ with features $F^{3D}$, 
       Scannet200 vocabulary $F^{vocab}$, points $S_O$, volumetric map $V$}
\Output{Updated reconstruction tasks $\mathcal{T}^{rec}$}

$O^{rec} \leftarrow RecInstanceGenerating(O, S_O, F^{3D}, F^{vocab})$\;
$U^{rec} \leftarrow InstanceUncertaintyRetrieving (U_{gs}, O^{rec})$\;

\ForEach{$O^{rec}_i \in O ^{rec}$}{
    \tcp{Process each instance}
    $S^d_i, U^d_i \leftarrow SurfaceDownsampling (S_i, U_i, N_{down})$\;
    $P^d_i \leftarrow POIRetrieving(S^d_i, U^d_i, d_{POI})$\;
    $VP_i^{rec} \leftarrow ViewSampling(V, S^d_i, U^d_i, P^d_i)$\;
    $v_i^{rec} \leftarrow ViewsofMaxGain (VP_i^{rec})$\;
    $\mathcal{T}_i^{rec} \leftarrow v_i^{rec}$\;
}

$\mathcal{T}^{rec} \leftarrow \{\mathcal{T}_1^{rec}, \mathcal{T}_2^{rec}, \ldots, \mathcal{T}_{N_{rec}}^{rec}\}$\;

\end{algorithm}
\vspace{-3mm}

\subsection{Task generation}
\seclabel{task_generation}
Within our approach, scanning tasks are categorized into two types: exploration tasks aimed at achieving rapid coverage and reconstruction tasks focused on ensuring high-quality reconstruction.

\subsubsection{Exploration task generation}
The exploration tasks $\mathcal{T}^{exp} = \{\mathcal{T}_1^{exp}, \mathcal{T}_2^{exp}, ..., \mathcal{T}_{N^{exp}}^{exp}\}$ are intended to cover more unknown regions, with $N^{exp}$ denoting the total number of tasks. We firstly update incremental frontiers and Euclidean signed distance field (ESDF) map $E$~\cite{han2019fiesta} from the maintained volumetric map $V=V_o \cup V_e \cup V_u$ similar to Fuel~\cite{zhou2021fuel}, where $V_o, V_e, V_u$ represent occupied, empty and unknown voxels. Subsequently, we select viewpoints that can provide superior coverage of the frontiers in exploration tasks, similar to our previous method AIISRFE~\cite{zeng2024autonomous}.

\subsubsection{Semantic-guided reconstruction task generation}
The reconstruction tasks $\mathcal{T}^{rec} = \{\mathcal{T}_1^{rec}, T_2^{rec}, \ldots,  \mathcal{T}_{N^{rec}}^{rec}\}$ are designed to refine areas with low reconstruction quality, and $N^{rec}$ denotes the total number of tasks. Generally, objects with complex textures and structures are more difficult to reconstruct, necessitating focused attention during scanning. Therefore, reconstruction tasks can be transformed into scanning the low-quality regions of objects with low completeness. Algorithm \ref{alg:rec} describes the generation process of reconstruction tasks. 

\noindent\textbf{Reconstruction instance generation} To evaluate the objectness scores~\cite{liu2018object} of 3D object instances $O$, which represents the probability that an instance belongs to a certain category and reflects the completeness of the object, we need a vocabulary to match each instance with its corresponding category. We choose ScanNet200 vocabulary~\cite{rozenberszki2022language} since it contains most indoor object categories. The similarity scores between the feature $f^{3D}_i$ of each instance $O_i \in O$ and ScanNet200 vocabulary features $F^{vocab}$ are computed and sorted in descending order as $\Lambda_i = \{\Lambda^1_i, \ldots, \Lambda^{200}_i\}$. To enhance inter-class separability, a scaling factor $\lambda_e = 50$ is applied to these scores. Finally, for each instance $O_i \in O$, the classification probabilities $P_i$ for the 200 categories are obtained by normalizing the scores using a softmax function as $C_i = \{C^1_i, \ldots, C^{200}_i\} = softmax(\lambda_e \Lambda_i)$. We select the category with the highest probability score as the label $L_i$ for instance $O_i$. The objectness score of this instance is defined as $C^1_i$.

If $C^1_i < C_{min}$, it is considered that the objectness score calculation of the instance may be affected by noise from CLIP labels. If $C^1_i > C_{max}$, it is deemed that the instance has already been well reconstructed. We exclude these two types of instances and denote the remaining instances as reconstruction instances $O^{rec}$:
\begin{equation}
O^{rec} = \{O^{rec}_1, \ldots, O^{rec}_m\}
\eqlabel{rec_instances}
\end{equation}
where $O^{rec}$ represents the instances used for generating reconstruction tasks, $m$ is the number of reconstruction instances. For each instance $O^{rec}_i = \{L^{rec}_i,C^{1,rec}_i,S^{rec}_i\} \in O^{rec}$, it includes the label $L^{rec}_i$, objectness score $C^{1,rec}_i$, and surface points $S^{rec}_i$, where $S^{rec}_i \in S_O$.

\noindent\textbf{Instance uncertainty} For each instance $O^{rec}_i \in O^{rec}$ in~\eqref{rec_instances}, we need to evaluate instance uncertainty to generate viewpoints that cover its low-quality areas. We query uncertainty of surface points $S^{rec}_i$ in instance $O^{rec}_i$ from surface uncertainty $U_{gs}$ and denotes it as instance uncertainty $U^{rec}_i$. The instance uncertainty for each instance within the reconstruction instances $O^{rec}$ can be obtained as follows
\begin{equation}
U^{rec} = \{U^{rec}_1, \ldots, U^{rec}_m\}
\eqlabel{instance_uncer}
\end{equation}

\noindent\textbf{View sampling} For each instance $O^{rec}_i \in O^{rec}$, we downsample its surface points by a factor of 
$N_{down}=5$, obtaining the downsampled surface points $S^d_i$ and uncertainty $U^d_i$. To generate task viewpoints, we need to identify the points of interest (POIs) for each instance. Specifically, we initially select the point with the highest uncertainty. Next, from the remaining points, we choose iteratively the point with the highest uncertainty that is at least $d_{POI}$ away from all previously selected points. This process continues until no more points can be added. The set of selected points is denoted as $P^d_i = \{P^d_{i,1}, \ldots, P^d_{i,k_i}\}$, where  
$k_i$ is the number of sampled POIs from instance $O^{rec}_i$. 

For each sampled POI $P^d_{i,j} \in P^d_i$, we set the center as $P^d_{i,j}$ and generate a series of candidate viewpoints $VP^d_{i,j}$ oriented towards the center within the empty space $Ve$, similar to AIISRFE~\cite{zeng2024autonomous}. This yields all candidate viewpoints $VP^d_i = \{VP^d_{i,1}, \ldots, VP^d_{i,k_i}\}$ for the instance $O^{rec}_i$.

\noindent\textbf{Instance uncertainty based Information gain} To select reconstruction tasks from these candidate viewpoints, we define information gain of viewpoint $v$ as:
\begin{equation}
g(v) = \sum_{k=1}^{N_{vis}} \sigma_k e^{-0.5d_{v,k}}
\end{equation}
where $N_{vis}$ represents the number of visible instance surface points, $\sigma_k \in U^d_i$ represents the uncertainty of each visible surface point $s_k \in S^d_i$, and $d_{v,k}$ is the distance from the viewpoint $v$ to the surface point $s_k$. 

We then choose the viewpoint $v^d_{i,j}$ with the highest information gain as the reconstruction task for the sampled POI $P^d_{i,j} \in P^d_i$ in instance $O^{rec}_i$. Reconstruction tasks $ \mathcal{T}_i^{rec} = v_i^{rec}$ for instance $O^{rec}_i$ are generated for all sampled POIs from instance $O^{rec}_i$. This allows us to obtain the final reconstruction tasks $\mathcal{T}^{rec} = \{\mathcal{T}_1^{rec}, \mathcal{T}_2^{rec}, \ldots, \mathcal{T}_{N_{rec}}^{rec}\}$.

\subsection{Hierarchical planning}
\seclabel{hierarchical_planning}
Once the new exploration tasks $\mathcal{T}^{exp}$ and reconstruction tasks 
$\mathcal{T}^{rec}$ are generated, they need to be assigned to the robots for execution. To find the best task assignment, we first construct a weighted graph $G^\mathcal{T}=(\mathcal{T}^{exp} \cup \mathcal{T}^{rec}, \mathcal{E})$ to represent the spatial relationships between the tasks and robots similar to ACAMS~\cite{guo2022asynchronous}, where $\mathcal{E}$ consists of edges connecting each pair of tasks with travel cost. We optimize the travel costs during task assignment and formulate it as an MDMTSP problem. The objective is to find a set of disjoint paths $\{\mathcal{T}^*_r\}^{N_R}_{r=1}$ that covers the entire set $G^\mathcal{T}$ such that the total tour costs for all robots are minimized:
\begin{equation}
E_d = \sum_{r=1}^{N_R} \sum_{\mathcal{T}_k \in \mathcal{T}^*_r} t_{lb}(\mathcal{T}_k, \mathcal{T}_{k+1})
\eqlabel{mdmtsp}
\end{equation}
where $N_R$ is the number of robots, $t_{lb}(\mathcal{T}_k, \mathcal{T}_{k+1})$ is the travel cost with the time lower bound from task $\mathcal{T}_k$ to $\mathcal{T}_{k+1}$ similar to Fuel~\cite{zhou2021fuel}, in which we consider path length cost, yaw cost and pitch cost. Moreover, directly merging exploration and reconstruction tasks is susceptible to falling into local minima, as demonstrated in~\cite{zeng2024autonomous}, making it essential to assign modes for the robots.

To solve the complex NP-hard MDMTSP problem as denoted in \eqref{mdmtsp}, we adopt a hierarchical planning strategy. In multi-robot collaboration, the robots are assigned to different modes and tasks to reduce the number of robot and task assignments in the MDMTSP problem. In single-robot planning, after the clusters with tasks are assigned to robots, the task execution order is determined for each robot.

\subsubsection{Multi-robot collaboration} In this section, we first determine the mode of each robot based on the global-local task distribution to reduce the number of robots and tasks in the MDMTSP problem. Subsequently, we employ an improved K-means clustering algorithm to assign tasks to the robots to reduce the number of task assignments within the problem.

\noindent\textbf{Mode assignment} The mode of the robot should be determined based on the distribution of nearby tasks due to the execution path cost. Besides, the number of robots assigned to exploration and reconstruction modes should also take into account the global task distribution, as demonstrated in~\cite{guo2022asynchronous}. To this end, we designed a global-local task distribution-guided method for robot mode assignment.

Firstly, we compute the proportion of two modes of all global tasks and assign the number of robots for each scanning mode according to this proportion.
Secondly, we improve the assignment efficiency through a local task count statistic to avoid the iterative optimization process in ACAMS~\cite{guo2022asynchronous}. Specifically, for each robot $R_r$, we count the number of exploration and reconstruction tasks $N^{exp}_{local,r}$, $N^{rec}_{local,r}$, within its local range not exceeding $d_{local}$. The overall score is denoted as $N^{ove}_{local,r} = N^{exp}_{local,r} - N^{rec}_{local,r}$. 
We then rank the robots based on their overall scores, with higher scores indicating a priority for exploration mode. This prioritization means that robots with more exploration tasks in their local area will be assigned to further exploration. Consequently, the modes for all robots are determined, considering the number of robots in each mode. 

\noindent\textbf{Task clustering and assignment} 
After assign different modes to robots, we need to assign tasks to robots within each mode. Although the number of robots and tasks is reduced through mode assignment, it is still an MDMTSP problem. To further reduce the complexity of solving this problem, we propose an improved K-means algorithm for task clustering and assignment.

For robots in task mode $M \in \{exp,rec\}$, we cluster tasks $\mathcal{T}^{M}$ into $N^{M}_R$ classes using an improved K-means method and design the following objective function for task clustering:
\begin{equation}
\min_{\mathcal{T}^M} \sum_{r=1}^{N^M_R} \underbrace{D_r + \gamma(R^M_r, \omega_r)}_{\text{moving cost}}+ \underbrace{\left( N^M_r - \overline{N} \right)^2 + |D_r - \overline{D}|}_{\text{robot capacity}}
\eqlabel{k-means}
\end{equation}
where $D_r = \sum_{\mathcal{T}_k \in \mathcal{T}^M_r} \gamma(\mathcal{T}_k, \omega_r)$ is the distance sum from assigned tasks to the task centroid $\omega_r$ for robot $R^M_r$ and $\overline{D} = \sum_{D_r} / N^M_R$ is the average distance of $N^M_R$ robots.  $\mathcal{T}^M_r \in \mathcal{T}^M$ is the tasks assigned to robot $R^{M}_r$. $\gamma(R^M_r, \omega_r)$ is the distance from robot $R^M_r$ to the task centroid $\omega_r$. $N^M_r$ is the number of tasks assigned to robot $R^{M}_r$ and $\overline{N} = N^M / N^M_R$ is average number of tasks for $N^M_R$ robots. Finally, we obtain the clusters of tasks assigned to all $N_R$ robots, denoted as $\{\mathcal{T}_r\}^{N_R}_{r=1}$. 

As above in~\eqref{k-means}, the task clustering and assignment are determined by the following two aspects:

\textit{Movement cost}. Similar to~\cite{dong2019multi}, We approximate the moving cost by considering two components: 1) the Euclidean distance from the tasks to the centroid of the set of assigned tasks, and 2) the Euclidean distance from the robot to the centroid of the set of assigned tasks.

\textit{Robot capacity}. Different from~\cite{dong2019multi, guo2022asynchronous}, to ensure load balancing constraints on different robots, we consider not only the balance of the number of tasks assigned to each robot but also the balance of the internal distances among the tasks assigned to each robot.

\subsubsection{Sing-robot view path planning}
In this section, we plan the view path for each robot with assigned tasks. For robot $R_r$ with assigned tasks $\mathcal{T}_r$, we plan a global shortest path based on ATSP similar to~\cite{zeng2024autonomous}
as follows:
\begin{equation}
\mathcal{T}^*_r = \mathop{\arg\min} \sum_{\mathcal{T}_k \in \mathcal{T}_r} t_{lb}(\mathcal{T}_k,\mathcal{T}_{k+1})
\eqlabel{ATSP}
\end{equation}
where $\mathcal{T}^*_r$ is planed path of robot $R_r$. For smooth navigation of the robot, we adopt B-spline optimization to generate smooth, safe, and dynamically feasible B-spline trajectories, similar to Fuel~\cite{zhou2021fuel}. Viewpoints $V^s_r$ are sampled along the trajectory at time intervals of $t=0.4s$, ensuring that the execution path length does not exceed $L_{exec}$. The sampled viewpoints for all robots are executed for map updates until a certain number of images are collected.

\section{Results}

\begin{table*}[!ht]
    \vspace{2mm}
    \centering
    \caption{Evaluations of the effectiveness and efficiency of view path for 3DGS and ASH representations.}
    \resizebox{\textwidth}{27 mm}{
    \normalsize
    \setlength{\tabcolsep}{1mm}{
    \begin{tabular}{c|cccc|cccc|cccc|cccc}
    \toprule
         \multicolumn{1}{c}{}& \multicolumn{4}{c}{Method} & \multicolumn{4}{c}{QaLdn} & \multicolumn{4}{c}{Nicut} & \multicolumn{4}{c}{Oyens}  \\ 
        Variant & Pitch & $\mathcal{T}^{rec}$ & MA & IKM & PSNR↑ & Acc↓ & Comp↓ & Recall↑ & PSNR↑ & Acc↓ & Comp↓ & Recall↑  & PSNR↑ & Acc↓ & Comp↓ & Recall↑ \\ 
        \midrule
        
        V1(Fuel swarm\textsuperscript{\cite{zhou2021fuel}}) & &  &  & & 19.30 & 10.15 & 2.89 & 0.87  & 18.10 & 3.06 & 2.51 & 0.90 & 18.71 & 3.54 & 4.65 &  0.81   \\ 
        V2 & \checkmark &  & & & 21.78 & 10.25 & 2.87 & 0.87 & 22.12 & 3.02 & 2.15 & 0.92 & 23.25 & 3.53 & 3.62 & 0.84 \\
        V3(AIISRFE swarm\textsuperscript{\cite{zeng2024autonomous}}) & \checkmark & Surface & & & 19.09 & 10.64 & 11.10 & 0.74  & 18.00 & 3.03 & 10.71 & 0.78 & 21.85 & 3.56 & 14.79 & 0.73 \\
        V4  & \checkmark & Semantics & &  & 20.33  & 10.35 & 5.38 & 0.79 & 21.51 & 3.03 & 3.43 & 0.89 & 22.56 & 3.52 & 10.24 & 0.75  \\
        V5  & \checkmark & Semantics & \checkmark &   & 21.89  & 10.25 & 2.41 & 0.89 & 22.84 & 3.05 & 1.91 & 0.94 & 23.35 & 3.54 & 4.17 & 0.83  \\
        V6(Ours full) & \checkmark & Semantics & \checkmark  & \checkmark & \textbf{22.75} & \textbf{10.04} & \textbf{2.25} & \textbf{0.91} & \textbf{24.28} & \textbf{3.01} & \textbf{1.62} & \textbf{0.95} & 
        \textbf{24.81} & \textbf{3.50} & \textbf{3.36} & \textbf{0.86}\\  
     
        \midrule
         Variant & Pitch & $\mathcal{T}^{rec}$ & MA & IKM & $T_{task}$  & $T_{colla}$ & $T_{GP}$ & P.L.  & $T_{task}$  & $T_{colla}$ & $T_{GP}$ & P.L. & $T_{task}$  & $T_{colla}$ & $T_{GP}$ & P.L.\\ \midrule
        V1(Fuel swarm\textsuperscript{\cite{zhou2021fuel}}) &  &  & & &  0.003  & 0.026 & 0.56 & 107.61 & 0.005 & 0.007 & 0.36 & 89.80 & 0.005 & 0.007 & 0.38 & 84.83\\
        V2 & \checkmark&  & & &  0.004  & 0.034 & 1.00 & 121.44 & 0.003 & 0.015 & 1.51 & 103.23 & 0.003 & 0.014 & 1.09 & 90.97\\
        V3(AIISRFE swarm\textsuperscript{\cite{zeng2024autonomous}})  & \checkmark & Surface & & & 2.735  & 0.110 & 18.77 & 98.10 & 2.232 & 0.103 & 18.53 & 93.39 & 2.785 & 0.550 & 18.44 & 78.68\\
        V4 &  \checkmark & Semantics & & &  0.799  & 0.353 & 10.15 & 97.19  &  0.477  & 0.109 & 4.12 & 107.95 & 0.432 & 0.231 & 5.03 & 108.60 \\
        V5  & \checkmark & Semantics & \checkmark &   & 0.701  & 0.241 & 9.16 & 144.49 & 0.421 & 0.147 & 4.10 & 129.46 & 0.441 & 0.263 & 5.56 & 113.33  \\
        V6(Ours full) & \checkmark & Semantics & \checkmark & \checkmark & 0.718  & 0.546 & 10.92 & 167.77 & 0.471  & 0.253 & 4.91 & 131.18 &
        0.437 & 0.280 & 5.85 & 126.46\\
     
         \midrule
    \end{tabular} 
    } 
    }
    \tablelabel{table_effect_effici}
     \vspace{-2mm}
\end{table*}

\begin{table*}[ht]
    \centering
    \caption{Evaluations of the effectiveness and efficiency with existing planning methods.}
    \resizebox{\textwidth}{11mm}{
    \normalsize
    \setlength{\tabcolsep}{1.0mm}{
    \begin{tabular}{c|cccccc|cccccc|cccccc}
    \toprule
        \multicolumn{1}{c}{} & \multicolumn{6}{c}{QaLdn} & \multicolumn{6}{c}{Nicut}&\multicolumn{6}{c}{Oyens}   \\ 
        Method  & PSNR↑ & Acc↓ & Comp↓ & Recall↑ & $T_{GP}$ & P.L. & PSNR↑ & Acc↓ & Comp↓ & Recall↑ & $T_{GP}$ & P.L. & PSNR↑ & Acc↓ & Comp↓ & Recall↑ & $T_{GP}$ & P.L. \\ 
        \midrule
        ACAMS\textsuperscript{\cite{guo2022asynchronous}} & 22.36 & 10.14 & 2.83 & 0.88 & 87.10 & 175.94 & 22.96 & 3.07 & 1.88 & 0.94 & 41.62 & 170.42 & 23.36 & 3.53 & 3.53 & 0.85 & 65.53 & 146.45\\ 
        
         MS3DSR\textsuperscript{\cite{hardouin2023multirobot}} & 21.05 & 10.19 & 2.96 & 0.87 & 24.87 & \textbf{121.90} & 20.75 & 3.09 & 2.46 & 0.91 & 13.62 & \textbf{120.16} & 19.80 & 3.53 & 25.67 & 0.65 & 8.78 & \textbf{125.51}\\ 
         
         Ours & \textbf{22.75}  & \textbf{10.04} & \textbf{2.25} & \textbf{0.91} & \textbf{10.92} & 167.77 & \textbf{24.28} & \textbf{3.01} & \textbf{1.62} & \textbf{0.95} & \textbf{4.91} & 131.18 & \textbf{24.81} & \textbf{3.50} & \textbf{3.36} & \textbf{0.86} & \textbf{5.85} & 126.46

          \\
        \midrule
    \end{tabular}
    }
    }
    \tablelabel{compare_with_existing_methods}
    \vspace{-4mm}
\end{table*}

\subsection{Implementation details}
\seclabel{impd}

\subsubsection{Data} The experiments are conducted on three virtual scenes: $QaLdn$ ($187m^2$) from HM3D~\cite{ramakrishnan2021hm3d}, $Nicut$ ($120m^2$) and $Oyens$ ($80m^2$) from Gibson~\cite{xiazamirhe2018gibsonenv}. These scenes are reconstructed from real-world environments. We maintain the same field of view parameters, as specified in prior works such as~\cite{zeng2023efficient,zeng2024autonomous}.

\subsubsection{Implementation} Our method runs on two GPUs. The 3DGS reconstruction and surface uncertainty updates are performed on one A6000 GPU; OVIR-3D and view path planning are on the other RTX3090 GPU, where view path planning runs in the ROS environment. We set the maximum planned views to be 600 views for $QaLdn$, 600 views for $Nicut$, and 500 views for $Oyens$. In reconstruction instance generation, objectness score thresholds are $C_{min}=0.2$, $C_{max}=0.6$. In view sampling, the POI distance threshold is $d_{POI} = 1.2m$. In mode assignment, the local range is $d_{local} = 6m$. In sing-robot planning, the execution path length is $L_{exec} = 6m$. The experiments are conducted using four robots.

\subsubsection{Metric} Similar to~\cite{zeng2023efficient,zeng2024autonomous}, we evaluate our method from two aspects including effectiveness and efficiency. The effectiveness is measured in two parts: the quality of the rendered images and the quality of the geometry of the reconstructed surface. We adopt metrics: Accuracy (cm), Completion (cm), Recall. The number of sampled points and completion distance threshold are the same as AIISRFE~\cite{zeng2024autonomous}.

For efficiency, we evaluate the path length (meter) and the planning time (second). The total path length is denoted as $P.L.$ and the planning time as $T_{GP}$. For a more detailed comparison, we further divide the time required for view path planning at each step of the reconstruction process into three components: 1) the task generation time $T_{task}$, 2) the time $T_{colla}$ for the multi-robot collaboration, 3) the time $T_{sr}$ for single-robot planning. $T_{sr}$ for all the variants in ~\tableref{table_effect_effici} is about 0.04s to 0.4s.


\begin{figure*}[htbp]
      \vspace{2mm}
      \centering
      \includegraphics[width=0.82\linewidth, height=0.27\linewidth]{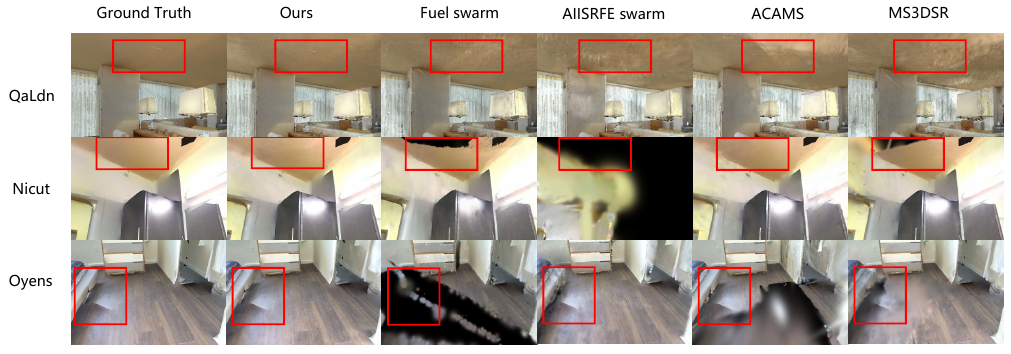}
    \includegraphics[trim=0cm 0cm 0cm 0.0cm, clip, width=0.8\linewidth,height=0.27\linewidth]{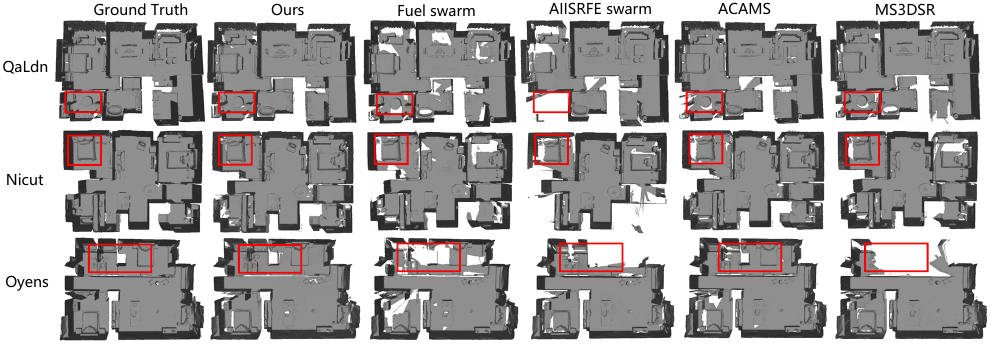}
      \caption{Comparison with different methods. Top: novel view synthesis from 3DGS; Bottom: reconstructed meshes from ASH.}
      \figlabel{views_geometry_compare}
      \vspace{-3mm}
\end{figure*}

\begin{figure*}[htbp]
      \centering
      \includegraphics[width=0.95\linewidth, height=0.15\linewidth]{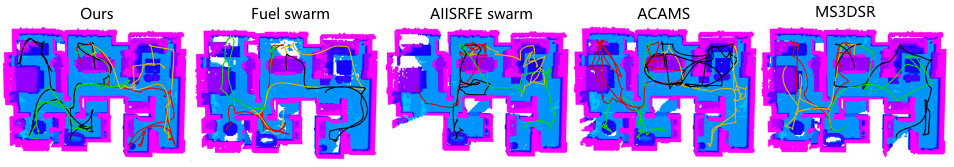}
      \caption{Comparison of trajectories with different methods.}
      \figlabel{trajectory_compare}
\vspace{-5mm}
\end{figure*}

\subsection{Efficacy of the Method} 
Similar to~\cite{zeng2024autonomous}, the efficacy of the method is evaluated regarding both the effectiveness and efficiency of our contributions. We design variants V1-V5 of our method based on 3DGS representation. V1 (Fuel swarm~\cite{zhou2021fuel}) considers only exploration tasks and extends Fuel to the multi-robot system. V2 adds a degree of pitch from V1. V3 (AIISRFE swarm~\cite{zeng2024autonomous}) considers exploration tasks and surface uncertainty-based reconstruction tasks and extends AIISRFE to the multi-robot system. V4 merges exploration and semantic-guided reconstruction tasks for assignment. V5 considers mode assignments for robots but uses moving cost and robot capacity in~\cite{dong2019multi} for task clustering and assignment. Our method is best for effectiveness and better than the method with surface-based reconstruction tasks for efficiency.

\subsubsection{Combination of exploration and semantic-guided reconstruction tasks} We make V1 and V2 as our baselines to verify the efficacy. To verify the efficacy of different reconstruction tasks, we make V2, V3, and V4 as our baselines. 

The metrics for V2 in Table I show that the pitch angle enhances reconstruction quality compared to Fuel swarm, with only a minor increase in planning time. Concentrating exclusively on exploration tasks (V2) neglects the detailed scanning of intricate features, which in turn diminishes the reconstruction quality. When exploration and surface-based reconstruction tasks (V3) are accounted for, the continuous updating of the KD-tree during task generation, coupled with the generation of numerous redundant viewpoints due to surface uncertainty error, further diminishes planning efficiency. The reconstruction quality becomes very poor because it falls into the local optimum and cannot cover the entire scene. V4 reduces task generation time due to efficiency of reconstruction instance generation.

\subsubsection{Multi-robot collaboration} To validate the effectiveness of mode assignment (\textbf{MA}) and improved K-means (\textbf{IKM}) based task assignment of robots, we use V4 and V5 as our baselines. The merger of these exploration and reconstruction tasks (V4) significantly slows the pace of scene exploration and can even result in getting trapped in a local optimum, thereby reducing the reconstruction quality. V5 does not consider balancing internal distances in \eqref{k-means}, which can result in long execution paths for some robots. Due to the limited energy of robots, path truncation is required for execution, which reduces exploration efficiency and, consequently, the quality of the reconstruction. (Ours) ensures the speed of exploration and detailed scanning while avoiding the local optima often encountered when scanning areas with high instance uncertainty.

\subsection{Comparison with existing planning methods} We select two recent works on multi-robot autonomous reconstruction: ACAMS~\cite{guo2022asynchronous} and MS3DSR~\cite{hardouin2023multirobot}, as our baselines. ACAMS proposes a modified MDMTSP and corresponding approximate solver to optimize each robot’s task assignment and execution order. MS3DSR generates viewpoints-based surface elements and introduces multiagent NBV planners to route robots to viewpoint configurations. The metrics presented in Table II indicate that our method excels compared to these baselines, showing superior reconstruction quality and planning efficiency. This is because the two aforementioned methods require continuous computation of the TSP when assigning modes and tasks, which increases time costs. Additionally, MS3DSR based on surface elements is susceptible to environmental factors such as holes, making it challenging to achieve rapid scene coverage and thereby reducing reconstruction quality. 

\figref{views_geometry_compare} shows our method provides better reconstruction results in novel views and geometry. For more visual comparisons and results, we refer readers to the supplementary video.~\figref{trajectory_compare} demonstrates that the trajectory of our method expands in scene $QaLdn$ that of other methods.

\subsection{Robot experiments in real scene}
We implement our proposed method on two Turtlebots equipped with an Azure Kinect camera\footnote{https://azure.microsoft.com/services/kinect-dk/} and a Realsense T265 camera\footnote{https://www.intelrealsense.com/tracking-camera-t265/} to perform room reconstruction, specifically targeting an exhibition room with dimensions of 12m × 8m × 4m. The pose of the Kinect camera is provided by the T265 camera. For this scene, turtlebots take about 8 minutes to explore and reconstruct the room. The exploration and reconstruction results will be presented in the supplementary video.

\section{Conclusion}

In this paper, we propose the first centralized multi-robot autonomous
3D reconstruction framework utilizing 3DGS. Subsequently, we incorporate online open-vocabulary semantic segmentation and surface uncertainty of 3DGS, focusing view sampling on areas with high instance uncertainty to reduce the time cost of task generation. Finally, we propose an efficient multi-robot collaboration strategy to improve reconstruction quality while ensuring planning efficiency. Comprehensive experiments demonstrate the superior performance of our method. 

In the future, we plan to investigate the deployment of our research on the distributed system to alleviate the communication load and enhance system robustness.

{\small
\bibliographystyle{ieeefullname}
\bibliography{egbib}
}

\end{document}